%% file: main.tex
\pdfoutput=1

\documentclass[11pt]{article}

\usepackage{acl}

\usepackage{times}
\usepackage{latexsym}
\usepackage{inconsolata}
\usepackage{enumitem}
\usepackage{amssymb}
\usepackage{mleftright}
\usepackage{xspace}

\usepackage[T1]{fontenc}

\usepackage[utf8]{inputenc}
\usepackage[font=footnotesize]{caption}

\usepackage{microtype}

\usepackage{graphicx}
\usepackage{bbm}
\usepackage{amsmath}
\usepackage{amsfonts}
\usepackage{amssymb}
\usepackage{amsthm}
\usepackage{dsfont}
\usepackage{caption}
\usepackage{subcaption}
\usepackage{xcolor}
\usepackage{soul}
\usepackage{hyperref}
\usepackage{adjustbox}
\usepackage{booktabs}
\usepackage{makecell}
\usepackage{nicefrac}
\usepackage{svg}
\usepackage{lipsum}
\usepackage{thmtools} 
\usepackage{thm-restate}
\usepackage{scalerel}
\usepackage{float}
\usepackage{flafter}
\usepackage{relsize}
\usepackage{mathtools}

\usepackage{thm-restate}

\usepackage[capitalize,noabbrev]{cleveref}
\crefname{section}{\S}{\S\S}
\crefname{table}{Tab.}{Tab.}
\crefname{figure}{Fig.}{Figs.}
\crefname{algorithm}{Alg.}{}
\crefname{equation}{Eq.}{Eq.}
\crefname{appendix}{App.}{}
\crefname{theorem}{Theorem}{}
\crefname{restatableTheorem}{Theorem}{}
\crefname{prop}{Proposition}{}
\crefname{definition}{Def.}{}
\crefname{cor}{Corollary}{}
\crefname{observation}{Observation}{}
\crefname{assumption}{Assumption}{}
\crefname{hyp}{Hyp.}{Hypotheses}
\crefformat{section}{\S#2#1#3}
\crefname{namedtheorem}{Hyp.}{Hypotheses}

\newtheorem{definition}{Definition}

\input{macros}

\usepackage[textsize=scriptsize,disable]{todonotes}

\usepackage{colortbl}
\definecolor{ETHBlue}{RGB}{33,92,175}   %
\definecolor{ETHGreen}{RGB}{98,115,19}      %
\definecolor{ETHPurpleDark}{RGB}{140,10,89} %
\definecolor{ETHPurple}{RGB}{163,7,116} %
\definecolor{ETHGray}{RGB}{111,111,111} %
\definecolor{ETHRed}{RGB}{183,53,45}    %
\definecolor{ETHPetrol}{RGB}{0,120,148} %
\definecolor{ETHBronze}{RGB}{142,103,19}    %
\definecolor{ETHOrange}{RGB}{230, 100, 50}

\colorlet{MacroColor}{ETHPetrol}
\colorlet{MACROCOLOR}{MacroColor}

\usepackage{fp}  %

\usepackage{tikz}
\usetikzlibrary{calc}

\newcommand{\surprise}[2]{%
	{\tt\tikz[baseline=(char.base),overlay]{
			\node[anchor=base] (char) {{\strut#1}}; %
			\node[above=-0.2em of char, fill=gray!20, minimum width=0.5em, minimum height={\dimexpr #2em / 4 \relax}, inner sep=0pt, outer sep=0pt] {}; %
		}\hspace{.5em}}%
}

\title{%
	\vspace{5ex}
	\surprise{O}{ 3.6540147461454957 }%
	\surprise{n}{ 1.09744848660154 }%
	\surprise{{\textvisiblespace}}{ 0.9145754323502189 }%
	\surprise{t}{ 1.5652420066263222 }%
	\surprise{h}{ 0.036025267157144825 }%
	\surprise{e}{ 0.244881492894514 }%
	\surprise{{\textvisiblespace}}{ 0.013835702777792669 }%
	\surprise{P}{ 5.17173714464623 }%
	\surprise{r}{ 1.5648204387433609 }%
	\surprise{o}{ 0.8709558950291516 }%
	\surprise{p}{ 2.549292292543745 }%
	\surprise{e}{ 1.0559016426485286 }%
	\surprise{r}{ 0.019114507051934737 }%
	\surprise{{\textvisiblespace}}{ 2.04697324625279 }%
	\surprise{T}{ 2.686353510480359 }%
	\surprise{r}{ 0.8795584981426252 }%
	\surprise{e}{ 0.22976154116281222 }%
	\surprise{a}{ 0.008103296916630143 }%
	\surprise{t}{ 0.005614096377605904 }%
	\surprise{m}{ 0.01831419461978001 }%
	\surprise{e}{ 0.000009 }%
	\surprise{n}{ 0.000009232455926877492 }%
	\surprise{t}{ 0.00000023301839355838183 }%
	\surprise{{\textvisiblespace}}{ 0.009906974088277565 }%
	\surprise{o}{ 0.0609388739709118 }%
	\surprise{f}{ 0.0010715699707475324 }%
	\surprise{{\textvisiblespace}}{ 0.0026103626365561183 }%
	\surprise{T}{ 3.264292396460153 }%
	\surprise{o}{ 1.7613159001835506 }%
	\surprise{k}{ 6.097487861889746 }%
	\surprise{e}{ 0.3953062301918564 }%
	\surprise{n}{ 0.04435416944994586 }%
	\surprise{i}{ 3.543952203751566 }%
	\surprise{z}{ 0.2655615238866389 }%
	\surprise{a}{ 1.6318451548922681 }%
	\surprise{t}{ 0.013222135081463193 }%
	\surprise{i}{ 0.0002781780552112423 }%
	\surprise{o}{ 0.000027558622662127163 }%
	\surprise{n}{ 0.000019517440222216464 }%
	\surprise{{\textvisiblespace}}{ 0.6206387166220679 }%
	\surprise{i}{ 1.6254519342700888 }%
	\surprise{n}{ 0.005853681210894024 }%
	\surprise{{\textvisiblespace}}{ 0.005698174195941874 }%
	\surprise{P}{ 2.3551946791224907 }%
	\surprise{s}{ 6.382647957574854 }%
	\surprise{y}{ 0.32399278326602854 }%
	\surprise{c}{ 0.050419058841697506 }%
	\surprise{h}{ 0.01980819270385581 }%
	\surprise{o}{ 0.5544155356555223 }%
	\surprise{l}{ 0.279937540255645 }%
	\surprise{i}{ 4.307810908783068 }%
	\surprise{n}{ 0.017205699118044038 }%
	\surprise{g}{ 0.0034113555734336387 }%
	\surprise{u}{ 0.00012981599281403078 }%
	\surprise{i}{ 0.00014636334012863017 }%
	\surprise{s}{ 0.00010113659268284891 }%
	\surprise{t}{ 0.000335221637520533 }%
	\surprise{i}{ 0.0006626452898927937 }%
	\surprise{c}{ 0.0000037862567992874574 }%
	\surprise{s}{0.2504424635481257}\hspace{-0.2em}\thanks{The gray bars above each character of the title are proportional to its character-level surprisal under {\tt GPT-2}.}
}

\newcommand{\person}[2]{%
	\ensuremath{\underset{\textnormal{\texttt{\href{mailto:#2}{#2}}}}{\textbf{#1}}}%
}

\author{%
	\person{Mario Giulianelli}{mario.giulianelli@inf.ethz.ch}~\;~\;~
	\person{Luca Malagutti}{luca.malagutti@inf.ethz.ch}~\;~\;~
	\person{Juan Luis Gastaldi}{juan.luis.gastaldi@inf.ethz.ch} \\
	\person{Brian DuSell}{brian.dusell@inf.ethz.ch}~\;~\;~
	\person{Tim Vieira}{tim.f.vieira@gmail.com}~\;~\;~
	\person{Ryan Cotterell}{rcotterell@inf.ethz.ch}\\
	\setlength{\fboxsep}{2.5pt}%
	\setlength{\fboxrule}{2.5pt}%
	\fcolorbox{white}{white}{%
		\includegraphics[width=2.5cm]{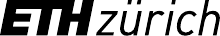}
	}
}

\begin{document}
	\maketitle
	
\begin{abstract}
	Language models are widely used in computational psycholinguistics to test theories that relate the negative log probability (the surprisal) of a region of interest (a substring of characters) under a language model to its cognitive cost experienced by readers, as operationalized, for example, by gaze duration on the region.
	However, the application of modern language models to psycholinguistic studies is complicated by the practice of using tokenization as an intermediate step in training a model.
	Doing so results in a language model over \emph{token} strings rather than one over character strings.
	Vexingly, regions of interest are generally misaligned with these token strings.
	The paper argues that token-level language models should be (approximately) marginalized into character-level language models before they are used in psycholinguistic studies to compute the surprisal of a region of interest; then, the marginalized character-level language model can be used to compute the surprisal of an arbitrary character substring, which we term a focal area, that the experimenter may wish to use as a predictor.
	Our proposal of marginalizing a token-level model into a character-level one solves this misalignment issue independently of the tokenization scheme.
	Empirically, we discover various focal areas whose surprisal is a better psychometric predictor than the surprisal of the region of interest itself.\looseness=-1
	
	\vspace{.11em}
	\hspace{1.25em}\includegraphics[width=1.25em,height=1.25em]{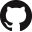}{\hspace{.75em}\parbox{\dimexpr\linewidth-2\fboxsep-2\fboxrule}
		{\url{https://github.com/rycolab/psycho-toke}}}
	
\end{abstract}

\section{Introduction}\label{sec:introduction}
Language models (LMs) have become a popular tool for computational psycho- and neurolinguists, who use them to instantiate and test executable linguistic theories~\citep{futrell2019neural,schrimpf2021,baroni2022proper}.
While there are various ways to operationalize theories of language processing using LMs \cite[e.g.,][]{caucheteux2023evidence,hoover2023plausibility,giulianelli-etal-2023-information,giulianelli-etal-2024-generalized,giulianelli-etal-2024-incremental,frank_2024}, their most common use in this area is to produce a specific quantity of interest: the probability of a character string given a context, conventionally mapped to the string's negative log probability, also known as its surprisal. 
Surprisal has been posited to correlate with the difficulty incurred by a comprehender processing that word~\cite{hale2001probabilistic,levy-2008}, and estimates of surprisal obtained from neural LMs have proven to be significant predictors of a broad range of psycholinguistic measurements of processing difficulty~\cite[][\textit{inter alia}]{goodkind2018predictive,shain2020fmri,wilcox-2023-testing,michaelov2024n400}, providing ample empirical support for the role of surprisal in psycholinguistics theory, in addition to other predictors, e.g., word length and unigram surprisal.

Modern language models do not provide direct access to contextual probabilities at the character level.
Instead, they provide a distribution over strings of \emph{tokens}, supercharacter units that are induced during a pre-processing step, e.g., by the byte-pair encoding tokenizer \citep[BPE;][]{sennrich-etal-2016-neural}.
However, in computational psycholinguistics, it is often necessary to compute the surprisal of an arbitrary character substring of the stimulus.
For example, the predictability of the first three characters of a word, which can be viewed parafoveally, is known to be an important predictor of whether the region is going to be skipped by the reader \citep[][\textit{inter alia}]{rayner1982availability,blanchard-etal-1989-acquisition,rayner2011eye}.
Therefore, to properly model a region's skip rate under surprisal theory, we should use the surprisal of the first three characters, rather than the surprisal of the entire region, as a predictor.
A complication arises when the first three characters of the region do not correspond to a token, and we have to marginalize over all token strings that start with those three characters.

Because performing such a marginalization over token strings is computationally expensive and therefore requires approximation \citep{cao-rimell-2021-evaluate,chirkova-etal-2023-marginalize,vieira-2024-token-to-char-lm}, it has yet to be adopted among computational psycholinguists.
Indeed, the literature lacks guiding principles for how computational psycholinguists should properly apply token-level language models to the field's inherently character-level problems.
For instance, a number of recent studies have blurred the line between algorithmic and linguistic concerns \citep[]
[\textit{inter alia}]{oh-etal-2021-surprisal,nair-resnik-2023-words,giulianelli-etal-2022-construction,beinborn-pinter-2023-analyzing,oh2024leading,pimentel2024compute,yee-etal-2024-efficiency}.
In particular, two recent studies \citep{oh2024leading,pimentel2024compute} suggest it is important to include a word's trailing whitespace in the computation of the word's surprisal to account for the mismatch between tokens and words.\footnote{See \citet{marantz2001words}, \citet{Haspelmath-2007-indeterminacy}, and \citet{krauska2023moving} for a discussion of various difficulties in defining the notion of a word in linguistic theory; thus, in the remainder of the paper, we use the term region of interest.} 
Both are motivated by a peculiarity\footnote{\label{fnlabel}The peculiarity in question is that standard implementations of BPE have the property that whitespace may occur at the beginning of a token but at no other place in the token.\looseness=-1} of the BPE tokenizer itself rather than a deeper appeal to linguistic theory: Examining standard practice in experimental eye-tracking research reveals that regions of interest are typically defined to include the \emph{preceding} whitespace rather than the trailing one \citep[][\textit{inter alia}]{rayner-1979-eye,pollatsek-1982-eye,mcconkie1987}.
We attribute the focus on trailing whitespace to the obfuscated relationship between token-level and character-level surprisal.\looseness=-1

Our paper clarifies the proper role of tokenization in surprisal theory: We take the stance that tokenization is irrelevant.
First, we note psycholinguistic stimuli should be viewed as \emph{character} strings rather than \emph{token} strings. 
This follows from the observation that linguists construct psycholinguistic stimuli without regard for any given LM's token alphabet. 
Then, as is common, the stimuli are divided into regions of interest (character substrings of the stimulus) for which the experimenter gathers a psycholinguistic measurement.
Note that treating regions of interest as character strings does not prevent the experimenter from claiming they represent morphemes, words, or phrases, all of which are built from characters in text form.
Finally, the experimenter collects the psycholinguistic measurements associated with each region.
None of the above steps makes use of tokenization schemes, completing our argument.\looseness=-1

A problem with respect to tokenization first arises when the experimenter \emph{analyzes} the measurements they collected by means of a language model and the regions of interest they decomposed their stimuli into do not neatly align with a string of tokens, or when the experimenter wishes to compute the surprisal of a sub- or super-string of the region.  
In this paper, we contend that the solution to this problem is to marginalize the token-level LM into a character-level one before using it to compute surprisal.
Moreover, on this view, such marginalization constitutes an algorithmic problem, but not a theoretical one for psycholinguistics.

Because computational psycholinguists have yet to convert pretrained token-level LMs into character-level ones through marginalization, we contend that they have yet to explore many potentially effective surprisal-based predictors.
In our experimental section, we test the degree to which the choice of various substrings of the stimulus that overlaps the regions of interest, which we term \emph{focal areas}, affect recent empirical findings in surprisal theory.
We perform such an exploration by means of \citeposs{vieira-2024-token-to-char-lm} approximate marginalization scheme. 
Across four datasets of eye-tracked reading times, we consistently find that computing the surprisal of the entire region of interest (with leading or trailing whitespace) rarely leads to the most effective surprisal-based predictor. 
For instance, as hinted at above, we observe that on the CELER dataset \citep{celer2022}, the surprisal of the first three characters is a significantly better predictor of skip rate.
On the Provo and MECO datasets \cite{provo-2018,siegelman2022expanding}, the surprisal of the first characters of a region---either determined by a fixed-size or a dynamically sized focal area extending over typical rightward word identification spans---is on par with the surprisal of the full region.
Finally, on the UCL dataset \citep{frank2013reading}, including a look-ahead focal area that peeks at the subsequent region significantly improves reading time predictions.\looseness=-1   

\section{Formalizing Psycholinguistic Stimuli} \label{sec:formalizing-stimuli}
We now offer an abstract formalization of the stimuli present in a sentence processing experiment.\looseness=-1

\subsection{Alphabets and Strings}
We overview the building blocks of digitalized language: alphabets and strings.
An \defn{alphabet} is a finite, non-empty set.
We use capital Greek letters to denote alphabets, e.g., we use $\charAlphabet$ for an alphabet of \defn{characters}, typically bytes.
Let $\kleene{\charAlphabet}$ be $\charAlphabet$'s Kleene closure, i.e., the set of all strings formed from $\charAlphabet$, and let $\kleeneplus{\alphabet} \defeq \kleene{\charAlphabet} \setminus \{\varepsilon\}$, where $\varepsilon$ is the empty string.
Given a string $\stimulus = \charSym{1} \cdots \charSym{N}$ of $N$ characters, we write $\stimulus_{(i, j)} = \charSym{i+1}\cdots \charSym{j-1}$, $\stimulus_{[i, j)} = \charSym{i}\cdots \charSym{j-1}$, $\stimulus_{(i, j]} = \charSym{i+1}\cdots \charSym{j}$ and $\stimulus_{[i, j]} = \charSym{i}\cdots \charSym{j}$ for $1 \le i \le j \le N$.
Furthermore, we write $\stimulus \preceq \stimulus''$ if $\stimulus$ is a prefix of $\stimulus''$ and $\stimulus \prec \stimulus''$ if $\stimulus$ is a proper prefix of $\stimulus''$.
We denote string concatenation with juxtaposition: $\stimulus\,\stimulus'$.\looseness=-1

\subsection{Regions of Interest}\label{sec:formalizing-roi}
In many psycholinguistic experiments, participants are presented with a character string $\stimulus$ as a \defn{stimulus}, and various neural or behavioral responses to the stimulus are measured.
To focus those measurements on parts of the string, the experimenter generally divides $\stimulus$ into regions of interest.
The most common regions of interest considered in psycholinguistic experiments are white-spaced separate substrings of the stimulus, often referred to as words.
However, one can just as easily experiment with regions of interest based on smaller (e.g., morphemes) or larger (e.g., constructions or sentences) units.
We now give a formal definition of a region of interest.\looseness=-1
\begin{definition}\label{def:roi}
	Let $\stimulus \in \charStrings$ be a non-empty stimulus. Let $N$ be its length.
	A \defn{region of interest} (\roi{}) $\region{} = [i, j)$ with $1 \leq i < j \leq N$ is a non-empty interval that correspond to a substring of $\stimulus$, which we denote as $\stimulus_{\region{k}}$.
	We refer to $\stimulus_{\region{k}}$ as the \roi{}'s \defn{yield} or simply the \roi{} when clear from context.
	We say that a sequence of regions of interest $\langle\region{k}\rangle_{k=1}^K$ is \defn{segmentative} if $\stimulus_{\region{1}}\cdots \stimulus_{\region{K}} = \stimulus$.\looseness=-1
\end{definition}

\subsubsection{Example 1: Self-paced Reading} \label{sec:example-self-paced}
In a self-paced reading experiment \cite{aaronson-1976-performance,just1982paradigms,rayner1998eye,enochson-2015-collecting}, the experimenter decomposes each stimulus $\stimulus \in \kleeneplus{\alphabet}$ into a sequence of \roi{}s $\langle\region{k}\rangle_{k=1}^K$.
The stimulus is then presented to the participant one \roi{} at a time, and the participant must click a button to progress to the next \roi{}.
The measurement associated with each \roi{} is the time elapsed between the initial presentation of an \roi{} and the participant's pressing of a button.\looseness=-1

As an example, consider the following stimulus taken from the UCL corpus \citep{frank2013reading}:\looseness=-1

\ex.\label{ex:sentence-reading-exp} Anne lost control and laughed.

When viewed as a string of characters $\stimulus \in \kleeneplus{\alphabet}$ (where $\alphabet$ is the set of Unicode symbols), \Cref{ex:sentence-reading-exp} is best thought of as the following Unicode string:
\ex.\label{ex:stimulus-reading-exp} $\charfont{Anne\textvisiblespace{}lost\textvisiblespace{}control\textvisiblespace{}and\textvisiblespace{}laughed.}$

where we have visualized the whitespace \charfont{\textvisiblespace{}} for clarity.
While it may seem like a triviality at first blush, graphical markers of boundaries, e.g., whitespace, do play a role in reading behavior \cite{pollatsek-1982-eye}.
Moreover, the whitespace~\charfont{\textvisiblespace{}} becomes relevant when we extract a surprisal estimate from a language model, as discussed in \Cref{sec:surprisal}.
Consider the following natural first pass at a sequence of \roi{}s, and related substrings $\langle \stimulus_{\region{k}} \rangle_{k=1}^K$, for \Cref{ex:stimulus-reading-exp}:

\ex.\label{ex:rois-reading-exp} 
$\langle
\charfont{Anne},  
\charfont{lost}, 
\charfont{control},
\charfont{and},
\charfont{laughed} 
\rangle$

\noindent with whitespaces and the sentence-final period omitted to accommodate the self-paced paradigm.
In our terminology, such a sequence of \roi{}s is called non-segmentative.
However, a segmentative sequence of \roi{}s is generally desired, and the following two sequences are natural choices:

\ex. \label{ex:rois-reading-exp-leading} 
$\langle
{\charfont{Anne}},  
{\charfont{\textvisiblespace{}lost}}, 
{\charfont{\textvisiblespace{}control}},
{\charfont{\textvisiblespace{}and}},
{\charfont{\textvisiblespace{}laughed.}} 
\rangle$

\ex. \label{ex:rois-reading-exp-trailing}
$\langle 
\charfont{Anne\textvisiblespace{}},  
\charfont{lost\textvisiblespace{}}, 
\charfont{control\textvisiblespace{}},
\charfont{and\textvisiblespace{}},
\charfont{laughed.} 
\rangle$

Indeed, \citet{oh2024leading} take the stance that \Cref{ex:rois-reading-exp-trailing} is a better choice than \Cref{ex:rois-reading-exp-leading}.
Specifically, they argue that, in a self-paced reading experiment where the regions of interest are generated by splitting the stimulus $\stimulus$ on the whitespace symbols $\charfont{\textvisiblespace{}}$, the reader knows the character string displayed on the screen \emph{must} be followed by the whitespace symbol~$\charfont{\textvisiblespace{}}$.
We, however, contest this point.
We agree, of course, that in such a setup, the participant knows that the symbols displayed must be followed by a whitespace~$\charfont{\textvisiblespace{}}$.
However, the reader symmetrically knows that the character string was \emph{preceded} by a whitespace~$\charfont{\textvisiblespace{}}$, but the surprisal of this preceding whitespace is attributed to the previous \roi{}.
Thus, that the reader has knowledge of an \roi{}'s surrounding whitespace, as is endemic to the self-paced reading paradigm, neither implies that the surprisal of those whitespace symbols should be lumped in with the \roi{}'s surprisal nor gives us a reason to include the trailing whitespace and to omit the preceding whitespace if we are forced to choose.\looseness=-1

Building on this, one alternative is to include the relevant whitespace in \emph{all} regions of interest:\looseness=-1
\ex. \label{ex:rois-reading-exp-leading-trailing} $
\langle
\charfont{Anne\textvisiblespace{}},
\charfont{\textvisiblespace{}lost\textvisiblespace{}},
\charfont{\textvisiblespace{}control\textvisiblespace{}},
\charfont{\textvisiblespace{}and\textvisiblespace{}},
\charfont{\textvisiblespace{}laughed.}
\rangle$

\noindent 
This choice, however, leads to a non-segmentative sequence of \roi{}s.
Such a non-segmentative sequence is undesirable because it makes it impossible to cleanly divvy up the measurements among the \roi{}s due to the overlap among them.
Nevertheless, choosing a surprisal-based predictor in a manner that takes into account both the preceding \emph{and} trailing whitespace may be a good and useful idea.
To accommodate this, we augment the notion of an \roi{} with that of a focal area, discussed in \cref{sec:focal-areas}.\looseness=-1

\subsubsection{Example 2: Eye-tracked Reading} \label{sec:example-eye-tracked}
We now turn our attention to eye-tracked reading, another widely used psycholinguistic method for studying real-time sentence processing \cite{rayner1998eye,rayner2006eye,frank2013reading}. 
In an eye-tracked reading experiment, participants naturally read a stimulus $\stimulus$ displayed on a screen while a camera tracks their eye movements.
Unlike the self-paced reading task, where the experimenter's flexibility in defining different types of \roi{}s is limited by the task's \roi{}-by-\roi{} design, the eye-tracked reading paradigm does not provide an inherent notion of an \roi{}.
Suppose a participant is presented with the stimulus string in \cref{ex:stimulus-reading-exp}.
It would be a natural decision for the experimenter to design segmentative \roi{}s with yields:
\ex.\label{ex:rois-reading-verb-phrase} $
\langle
\charfont{Anne},
\charfont{\textvisiblespace{}lost\textvisiblespace{}control},
\charfont{\textvisiblespace{}and\textvisiblespace{}laughed.}
\rangle$

\noindent if they were interested in studying verb phrases, and to then allot the measurements accordingly.

More frequently, however, the experimenter would define a segmentative sequence of \roi{}s split around whitespace boundaries.
This might result in sequences of \roi{}s such as \cref{ex:rois-reading-exp-leading} and \cref{ex:rois-reading-exp-trailing}.
This choice of \roi{}s might be meaningful, for instance, for a study on fixation duration under the hypothesis that whitespaces aid word identification processes and should thus have an effect on saccade latency \citep{fisher-1975-reading,malt-season-1978-peripheral}.
An alternative to segmentative \roi{}s such as \cref{ex:rois-reading-exp-leading} and \cref{ex:rois-reading-exp-trailing} would be to exclude whitespaces from \roi{}s as in \cref{ex:rois-reading-exp}, with reading time measurements post-processed such that only fixations to the five whitespace-separated substrings composing the sentence are retained. 
This would be in line with evidence that readers use word boundary information in saccade planning to decide \textit{where}---rather than \textit{when}---to move their gaze \cite{rayner-pollatsek-1981-eye,pollatsek-1982-eye}, and that therefore whitespaces should not have an effect on saccade latency.
However, excluding fixations on whitespace would discard data.\looseness=-1

\subsection{Much Ado About Trailing Whitespace}\label{sec:much-ado}

Two recent studies \citep{oh2024leading,pimentel2024compute} suggest that it is important to include the trailing whitespace and exclude the preceding whitespace in the definition of \roi{}s\footnote{Both studies refer to \roi{}s as words.} in the context of surprisal theory; indeed, \citet{pimentel2024compute} state that the exclusion of a trailing whitespace is incorrect.
With this backdrop, we offer an alternative view on this prescription.
First and foremost, \roi{}s are typically determined by the experimenter who collected the dataset, and their choice of \roi{}s is already reflected in the psycholinguistic measurements reported; see \Cref{sec:data}.
Thus, whether or not we wish to post-pend a trailing whitespace to an \roi{}'s yield should primarily be informed by the data collection itself.
Luckily, the role of whitespace in eye-tracking studies is already heavily investigated. 
In the traditional eye-tracking paradigm, \roi{}s are typically defined such that their yields include the preceding whitespace \citep[][\textit{inter alia}]{rayner-1979-eye,pollatsek-1982-eye,mcconkie1987}.
The justification for this choice stems from the fact that ``people tend to direct their gaze
to a point just left of the center of a word more
frequently than to other locations'' \citep{mcconkie1987}. 
Moreover, as it is standard practice for \roi{}s to be segmentative (including whitespace in their yields) as the reader is taken to be fixating on one \roi{} at a time, the trailing whitespace must be excluded.

The situation is slightly different in the context of the digital corpora annotated with eye-tracked reading times used for larger-scale surprisal studies.
For instance, Provo \citep{provo-2018}, MECO \citep{siegelman2022expanding}, CELER \citep{celer2022}, and PoTeC \citep{potec} all divide the separating visual whitespaces equally between the preceding and the trailing \roi{}.
When the yields of the \roi{}s are presented to a participant on a screen, as is the case with eye-tracking studies, it is possible to divide the whitespace displayed on the screen in half, associating the fixations in each half to the corresponding \roi{}. 
However, in the context of character strings one cannot perform such a splitting: The whitespace symbol \charfont{\textvisiblespace{}} is indivisible. 
Thus, in the case of these corpora, that is up to the modeler to determine how they wish to associate the indivisible whitespace symbol \charfont{\textvisiblespace{}} with the \roi{}; the data cannot inform the decision.
In these cases, we view whether one includes a preceding or trailing whitespace in the yields of the \roi{} as an empirical question and not a matter of correctness. 
No eye-tracking study to the authors' knowledge, however, associates a trailing whitespace with an \roi{}.\looseness=-1 

Finally, we note there is no inherent linguistic reason a trailing whitespace belongs to its preceding \roi{}, but contend that the \roi{}s used in surprisal studies should be the same \roi{}s the experimenter collected the data under.
However, in \Cref{sec:focal-areas}, we discuss a more general abstraction for choosing what substring one should compute the surprisal of that is more detached from the specific choice of \roi{} in an attempt to resolve the tension between choosing the most useful surprisal-based predictors and respecting the data as they were collected.\looseness=-1

\subsection{Psycholinguistic Data}\label{sec:data}
Saccade latency, discussed at the end of \cref{sec:example-eye-tracked}, is one example of the type of data gathered in a psycholinguistic experiment.
More generally, during experimentation, we say a psycholinguist collects a \defn{measurement}, abstractly denoted $\datum(\region{k}) \in \R$ for each \roi{} $\region{k}$.
The measurement is typically a neural or behavioral response of the participant to consuming the stimulus segment $\region{k}$, e.g., the time elapsed between keystrokes in a self-paced reading experiment \citep{just1982paradigms}, the duration of a participant's first fixation on the \roi{} \citep{rayner1998eye}, or the voltages produced by neural activity corresponding to that fixation \citep{donchin1979event}.
To explain or better understand the measurements for each \roi{}, the psycholinguist constructs a set of explanatory variables to predict the measurements. 
Regression analysis is applied to gain insight into the underlying aspects of human language processing that generated the measurements.

\section{Language Models as Predictors}\label{sec:surprisal}
In psycholinguistics, LM-derived predictors are commonly used to predict measurements such as the participants' reading time for an \roi{}.
We now give an overview of the necessary background. 

\subsection{Language Modeling}
A \defn{language model} $\pLM$ is a probability distribution over $\kleene{\charAlphabet}$. We define $\pLM$'s \defn{prefix probability} $\pLMprefix(\charseq)$ as the probability that a string drawn from $\pLM$ begins with a particular string $\charseq \in \charStrings$:\looseness=-1
\begin{equation}\label{eq:prefix-prob}
	\pLMprefix(\stimulus) 
	\defeq 
	\smashoperator{\sum_{\stimulus' \in \kleene{\charAlphabet}}} \mathbbm{1}\mleft\{ \stimulus \preceq \stimulus' \mright\}\pLM(\stimulus')~.
\end{equation}
Prefix probabilities are primarily used to compute the conditional probability of the continuation $\stimulus' \in \charStrings$ given a preceding context $\stimulus$:\footnote{
	Note that $\pLMprefix(\varepsilon \mid \stimulus) = 1$ for all $\stimulus \in \kleene{\alphabet}$ and \cref{eq:ratio} is only well-defined when $\pLMprefix(\stimulus)>0$, a condition which will always be satisfied for softmax-normalized language models.\looseness=-1}
\begin{equation}\label{eq:ratio}
	\pLM(\stimulus' \mid \stimulus) 
	=
	\frac{\pLMprefix(\stimulus \c \stimulus')}{\pLMprefix(\stimulus)}.
\end{equation}
We can factorize a language model $\pLM$ as
\begin{equation}
	\label{eq:conditional-prob-product}
	\pLM(\stimulus) = \pLM(\eos \mid \stimulus) \prod_{t= 1}^{|\stimulus|} \pLM(\charSym{t} \mid \stimulus_{[1,t)}),
\end{equation}
where each $\pLM(\charSym{t} \mid \stimulus_{[1,t)})$ is a conditional probability over the set $\charAlphabet \cup \{\eos\}$, $\eos \not\in \charAlphabet$ is a distinguished end-of-string symbol, and\looseness=-1
\begin{equation}
	\label{eq:eos-prob}
	\pLM(\eos \mid \stimulus) \defeq \frac{\pLM(\stimulus)}{\pLMprefix(\stimulus)} .
\end{equation}

\paragraph{The human language model.}
Much work in computational psycholinguistics builds on the assumption that humans process language probabilistically, i.e., that humans have an internal language model. 
We denote the hypothetical construct of a human language model as $\pHum$.
Because the true human language model is unknown, we must approximate it via another language model, which we will call $\pLM$.
To the extent that $\pLM$ is a good approximation of $\pHum$, we would expect estimates derived from $\pLM$ to be a reliable proxy of the probabilities prescribed by the human language model.\looseness-1

\subsection{Surprisal Theory}
One popular information-theoretic framework for deriving computational predictors from language models is surprisal theory \citep{hale2001probabilistic,levy-2008}.
Surprisal theory states that the predictability of an \roi{}'s yield in the context of its preceding character string is a useful predictor for the \roi{}'s psycholinguistic measurements.
To define surprisal formally, we introduce additional notation.
Let $\stimulus \in \kleeneplus{\alphabet}$ be a stimulus divided into $K$ \roi{}s $\langle\region{k}\rangle_{k=1}^K$.
Then, the \defn{surprisal} of an \roi{} $\region{k} = [i, j)$ in context $\stimulus_{[1, i)}$ is\looseness=-1
\begin{equation} \label{eq:surprisal}
	\surprisal(\stimulus_{[i, j)} \mid \stimulus_{[1, i)}) \defeq - \log \pLM(\stimulus_{[i, j)} \mid \stimulus_{[1, i)}).
\end{equation}
We remark again on a key latent assumption embedded in surprisal theory: It is assumed that the language model $\pLM$ used to compute \Cref{eq:surprisal} well-approximates the human language model $\pHum$, as discussed above. 
Surprisal theory then posits that the surprisal of an \roi{} in context is a good predictor of measurements $\datum$ that seek to operationalize processing difficulty, e.g., reading time.
Empirically, this result has been demonstrated in many studies \citep[][\textit{inter alia}]{smith2013-log-reading-time,goodkind2018predictive,shain2020fmri,merkx-frank-2021-human,wilcox-2023-testing}.

\subsection{Focal Areas}\label{sec:focal-areas}
Computing the surprisal of the \roi{}'s entire yield in context is often too coarse-grained.
To allow for additional modeling freedom, we further associate every \roi{} with one or more focal areas, i.e., the portion of the \roi{}'s substring or the characters surrounding it to which the experimenter assigns a special status in terms of computing surprisal values.
We can regard a focal area as a string-valued feature of an \roi{} which tells us which surprisal to compute. 
Indeed, we take the stance that there is no inherent reason why the substring of the stimulus one computes the surprisal of should be identical to the \roi{}'s yield.\looseness=-1
\begin{definition}\label{def:focal-area}
	Let $\stimulus \in \charStrings$ be a non-empty stimulus. Let $N$ be its length, and let $\region{k} = [i, j)$ be a region of interest.
	Then, a \defn{focal area} of $\region{k}$ is a non-empty interval $[\alpha_k,\beta_k)$ with $i \leq \beta_k$,  $\alpha_k \leq j$ and $1 \leq \alpha_k < \beta_k \leq N$ corresponding to the substring $\stimulus_{[\alpha_k,\beta_k)}$ of the stimulus.
	We refer to $\stimulus_{[\alpha_k,\beta_k)}$ as the focal area's \defn{yield} or simply the focal area when clear from context.
\end{definition}

\begin{remark}
	\Cref{def:focal-area} states that a focal area must overlap with its corresponding \roi{}, i.e., have a non-empty intersection. 
	This constraints the focal area from being fully disassociated from the \roi{}.\looseness=-1
\end{remark}
\noindent The surprisal of \roi{} $\region{k}$'s focal area $[\alpha_k,\beta_k)$ is 
\begin{equation}\label{eq:surprisal-focal-area}
	\surprisal(\stimulus_{[\alpha, \beta)} \mid \stimulus_{[1, \alpha)}) \defeq - \log \pLM(\stimulus_{[\alpha, \beta)} \mid \stimulus_{[1, \alpha)}).
\end{equation}
Focal areas allow the modeler to express that, in some circumstances, the psycholinguist hypothesizes, or assumes, that the non-focal areas of the region will not have an influence on the measurements collected for that \roi{}, or that characters outside of the \roi{}'s yield will.\footnote{Indeed, the experimenter may devise multiple focal areas they believe have distinct influences on the measurements.}

\paragraph{Example: Modeling skipped \roi{}s.}
To understand the utility of focal areas, consider an eye-tracked reading experiment where the measurement of interest for any given \roi{} is its skip rate, i.e., the proportion of experimental trials in which the participant did not fixate on that \roi{} \cite{rayner2011eye}.
The experimenter might design a stimulus such as \cref{ex:stimulus-reading-exp} and the respective segmentative sequence of \roi{}s with trailing whitespaces, as in \cref{ex:rois-reading-exp-trailing}.
A reader's decision to skip a target region, say $\stimulus_{\region{3}} = \charfont{\charfont{control\textvisiblespace{}}}$, cannot be made when fixating on the entire ROI---otherwise, we could not say that $\charfont{\charfont{control\textvisiblespace{}}}$ had been skipped.
Instead, the decision must be made when fixating on characters preceding the region.
However, when fixating on the preceding characters, the reader has partial access to characters to the right \cite{rayner1982availability,underwood-mcconkie-1985-perceptual,mcconkie-zola-1987-visual}, which belong to the subsequent \roi{}.
Thus, the decision to skip the target region should depend at most on the first few characters of the \roi{}'s yield.
To determine the exact number of characters in the focal area, the psycholinguist may build on prior empirical evidence (see \cref{sec:selecting-focal-areas}).
For example, \citet{rayner1982availability} found that when the first three characters of the \roi{}'s yield to the right of the fixation were available, and the remainder of the characters were replaced, the reading rate was not substantially affected.
In line with such evidence, the psycholinguist may design a focal area on $\stimulus_{[11, 14)} = \focalareafont{con}$ consisting of the first three characters of $\charfont{\focalareafont{con}trol\textvisiblespace}$, the yield of \roi{} $\region{3}$.
The skip rate for $\region{3}$ would then be modeled using the surprisal of the focal area, $\surprisal(\focalareafont{con} \mid \charfont{Anne\textvisiblespace{}lost\textvisiblespace{}})$, as a predictor.\looseness=-1

\paragraph{The role of $\eos$ in focal areas.}
Because $\eos \not\in \alphabet$, by the definition of a stimulus as a character string and \roi{}s as intervals corresponding to substrings of the stimulus, $\eos$ cannot be included in an $\roi{}$'s yield.
However, when analyzing wrap-up effects \citep{meister-etal-2022-analyzing}, it may be prudent to abuse the definition and include $\eos$ anyway.\looseness=-1

\begin{table*}[t]
	\centering
	\resizebox{\textwidth}{!}{
		\begin{tabular}{l|l|l}
			\toprule
			\textbf{}                          & \textbf{Leading Whitespace} & \textbf{Trailing Whitespace}  \\ \midrule
			Full \roi{} & $\langle \charfont{\textvisiblespace{}lost}$, $\charfont{\textvisiblespace{}control}$, $\charfont{\textvisiblespace{}and}$, $\charfont{\textvisiblespace{}laughed.}\rangle$ & $\langle \charfont{lost\textvisiblespace{}}$, $\charfont{control\textvisiblespace{}}$, $\charfont{and\textvisiblespace{}}$, $\charfont{laughed.}\rangle$\\
			Fixed & $\langle \charfont{\textvisiblespace{}lo}$, $\charfont{\textvisiblespace{}co}$, $\charfont{\textvisiblespace{}an}$, $\charfont{\textvisiblespace{}la}\rangle$ & $\langle \charfont{los}$, $\charfont{con}$, $\charfont{and}$, $\charfont{lau}\rangle$ \\
			Dynamic (7) & $\langle \charfont{\textvisiblespace{}lost}$, $\charfont{\textvisiblespace{}cont}$, $\charfont{\textvisiblespace{}an}$, $\charfont{\textvisiblespace{}laug}\rangle$ & $\langle \charfont{lost\textvisiblespace{}}$, $\charfont{contr}$, $\charfont{and}$, $\charfont{laugh}\rangle$\\
			Dynamic (8) & $\langle \charfont{\textvisiblespace{}lost}$, $\charfont{\textvisiblespace{}contr}$, $\charfont{\textvisiblespace{}and}$, $\charfont{\textvisiblespace{}laugh}\rangle$ & $\langle \charfont{lost\textvisiblespace{}}$, $\charfont{contro}$, $\charfont{and\textvisiblespace{}}$, $\charfont{laughe}\rangle$\\
			Look-ahead (3) & $\langle \charfont{\textvisiblespace{}lost\textvisiblespace{}co}$, $\charfont{\textvisiblespace{}control\textvisiblespace{}an}$, $\charfont{\textvisiblespace{}and\textvisiblespace{}la}$, $\charfont{\textvisiblespace{}laughed.}\rangle$ & $\langle \charfont{lost\textvisiblespace{}con}$, $\charfont{control\textvisiblespace{}and}$, $\charfont{and\textvisiblespace{}lau}$, $\charfont{laughed.}\rangle$\\
			Look-ahead (4) & $\langle \charfont{\textvisiblespace{}lost\textvisiblespace{}con}$, $\charfont{\textvisiblespace{}control\textvisiblespace{}and}$, $\charfont{\textvisiblespace{}and\textvisiblespace{}lau}$, $\charfont{\textvisiblespace{}laughed.}\rangle$ & $\langle \charfont{lost\textvisiblespace{}cont}$, $\charfont{control\textvisiblespace{}and\textvisiblespace{}}$, $\charfont{and\textvisiblespace{}laug}$, $\charfont{laughed.}\rangle$\\
			Look-ahead (5) & $\langle \charfont{\textvisiblespace{}lost\textvisiblespace{}cont}$, $\charfont{\textvisiblespace{}control\textvisiblespace{}and\textvisiblespace{}}$, $\charfont{\textvisiblespace{}and\textvisiblespace{}laug}$, $\charfont{\textvisiblespace{}laughed.}\rangle$ & $\langle \charfont{lost\textvisiblespace{}contr}$, $\charfont{control\textvisiblespace{}and\textvisiblespace{}l}$, $\charfont{and\textvisiblespace{}laugh}$, $\charfont{laughed.}\rangle$\\
			Look-ahead (6) & $\langle \charfont{\textvisiblespace{}lost\textvisiblespace{}contr}$, $\charfont{\textvisiblespace{}control\textvisiblespace{}and\textvisiblespace{}l}$, $\charfont{\textvisiblespace{}and\textvisiblespace{}laugh}$, $\charfont{\textvisiblespace{}laughed.}\rangle$ & $\langle \charfont{lost\textvisiblespace{}contro}$, $\charfont{control\textvisiblespace{}and\textvisiblespace{}la}$, $\charfont{and\textvisiblespace{}laughe}$, $\charfont{laughed.}\rangle$\\
			Look-ahead (7) & $\langle \charfont{\textvisiblespace{}lost\textvisiblespace{}contro}$, $\charfont{\textvisiblespace{}control\textvisiblespace{}and\textvisiblespace{}la}$, $\charfont{\textvisiblespace{}and\textvisiblespace{}laughe}$, $\charfont{\textvisiblespace{}laughed.}\rangle$ & $\langle \charfont{lost\textvisiblespace{}control}$, $\charfont{control\textvisiblespace{}and\textvisiblespace{}lau}$, $\charfont{and\textvisiblespace{}laughed}$, $\charfont{laughed.}\rangle$\\
			Look-ahead (Full) & $\langle \charfont{\textvisiblespace{}lost\textvisiblespace{}control}$, $\charfont{\textvisiblespace{}control\textvisiblespace{}and}$, $\charfont{\textvisiblespace{}and\textvisiblespace{}laughed.}$, $\charfont{\textvisiblespace{}laughed.}\rangle$ & $\langle \charfont{lost\textvisiblespace{}control\textvisiblespace{}}$, $\charfont{control\textvisiblespace{}and\textvisiblespace{}}$, $\charfont{and\textvisiblespace{}laughed.}$, $\charfont{laughed.} \rangle$\\ \bottomrule                 
		\end{tabular}%
	}
	\caption{Yields of the focal areas for the stimulus $\stimulus = \charfont{Anne\textvisiblespace{}lost\textvisiblespace{}control\textvisiblespace{}and\textvisiblespace{}laughed.}$ using two segmentative sequences of \roi{}s (see \cref{sec:formalizing-roi}) and ten focal areas (see \cref{sec:selecting-focal-areas}). 
		The first \roi{} is skipped to ensure all focal areas are well-defined.
	}
	\label{tab:focal-areas-example}
\end{table*}

\subsection{Selecting Focal Areas}\label{sec:selecting-focal-areas}
We now explain our construction of various focal areas based on insights from human language processing and the psychology of reading.\looseness=-1

\paragraph{Dynamically sized focal areas.}
The perceptual span during reading, which is the range of visual information available around the fixation point, is relatively limited for readers of alphabetical orthographies such as English.
It typically extends from about 3--4 characters to the left of the fixation point to approximately 14--15 characters to the right \cite{McConkie-rayner-1975-span,mcconkie-rayner-1976-asymmetry,rayner-bertera-1979-reading,rayner-etal-1981-masking,denbuurman-etal-1981-eye,underwood-mcconkie-1985-perceptual} for English readers.
However, the span within which words can actually be identified, known as the \defn{word identification span}, is narrower, generally extending no more than 7--8 characters to the right of the fixation \cite{rayner1982availability,mcconkie-zola-1987-visual,underwood-mcconkie-1985-perceptual}.
Furthermore, readers' preferred viewing location, i.e., the location where they typically land after a saccade, tends to be a character between the beginning and the middle of the \roi{} \citep{oregan-1980-control,rayner-1979-eye}, approximately at position $\lceil\nicefrac{|\stimulus_{\region{k}}|}{2} \rceil - 1$ for the $k^{\text{th}}$ ROI \cite{rayner-pollatsek-1981-eye}. 
Thus, the size of the focal area on the initial characters of the upcoming region $\region{k+1}$ should vary depending on the length of $\region{k}$.
With a \defn{preferred viewing location} on the character at position $\prefViewLoc = \lceil\nicefrac{|\stimulus_{\region{k}}|}{2} \rceil - 1$, and a \defn{rightward word identification span} of $\wordIdSpan \in \{7,8\}$
characters, the focal area for region $\region{k+1}$ should include the first $\min(|\stimulus_{\region{k+1}}|, \max(0, \prefViewLoc + \wordIdSpan - |\stimulus_{\region{k}}|))$ characters of the region.
See the rows labeled Dynamic in \Cref{tab:focal-areas-example} for an example.\looseness=-1

\paragraph{Fixed-size focal areas.} 
Alternatively, the design of a focal area could be fixed in size. 
Research indicates that the initial characters of parafoveal \roi{}s are crucial not merely due to their proximity to the fixation point but because they aid in initiating lexical access and integrating information across fixations \cite{inhoff-1989-lexical,inhoff-1990-integrating,inhoff-tousman-1990-lexical}. 
Multiple studies show that previewing exactly the first three characters of a word, even with the remaining characters replaced by visually similar ones, enhances reading speed \cite{rayner1982availability,lima-inhoff-1985-lexical,lima-1987-morphological}, and that parafoveal previews also allow readers to skip words up to three characters long \cite{blanchard-etal-1989-acquisition}. 
Consequently, a fixed-size focal area, consistently covering the first $\min(|\region{k}|, 3)$ characters of a region, might be an effective predictor for that \roi{}'s collected measurements.
See the row labeled Fixed in \Cref{tab:focal-areas-example} for an example.\looseness=-1

\paragraph{Look-ahead focal areas.}
Finally, we design a focal area that looks ahead, i.e., one that includes characters to the right of the \roi{}'s yield and into the next \roi{}'s yield.
The argument for designing a look-ahead focal area stems from the fact we may wish to model the structural integration cost that could arise if the \roi{} corresponds to the end of a constituent---or, symmetrically, the additional processing cost that could arise when creating a new constituent \citep{gibson2001dependency,futrell2020lossy}.
However, without look-ahead into the next \roi{}'s yield, it can be difficult to judge whether it is necessary to integrate a new constituent.
Finally, we remark that focal areas that admit look-ahead resolve the problem of how to associate whitespace with \roi{}s as they detether defining a sequence of \roi{}s from the surprisal computation (see \Cref{sec:much-ado}): A sequence of \roi{}s that incorporates preceding whitespace, to respect how the psycholinguistics measurements were collected, can still be associated with a surprisal value that includes the surprisal of that \roi{}'s trailing whitespace. 
Of course, whether including this trailing whitespace helps remains an empirical question. In our experiments, we use look-aheads of 3 to 7 characters as well as a look-ahead peeking into the entire upcoming \roi{}.
See the rows labeled Look-ahead in \Cref{tab:focal-areas-example} for an example.\looseness=-1

\paragraph{Focal areas in past studies.}
In most past studies, experiments predicting measurements of reading behavior typically compute the surprisal of the entire \roi{} without accounting for specific focal areas \citep[][\textit{inter alia}]{goodkind2018predictive,wilcox2020,wilcox-2023-testing,shainetal24}. 
Most of these studies use LMs that rely on the BPE tokenizer and, thus, the \roi{}s' yields are defined to include the \emph{preceding} whitespace due to an oddity of BPE (see Footnote \ref{fnlabel}), but recall \Cref{sec:much-ado} for two exceptions.\looseness=-1

\section{Marginalizing Out Token Strings}\label{sec:marginalizing-out-token-strings}
The previous two sections (\cref{sec:formalizing-stimuli} and \cref{sec:surprisal}) have formalized psycholinguistic stimuli, their regions of interest, and focal areas at the character level.
Indeed, in this discussion, the character-level model used to compute surprisal is agnostic as to whether the model underlyingly uses tokenization or not. 

However, tokenization has evolved into a standard practice in constructing language models.
Rather than constituting a distribution over $\kleene{\alphabet}$, the set of all character strings, most modern language models are distributions $\ptokens$ over $\tokStrings$ where $\tokAlphabet$ is an alphabet of \defn{tokens}.  
To \defn{encode} a character string as a token string, we apply a function of type $\tokenizer\colon \charStrings \to \tokStrings$.  To \defn{decode} a token string to a character string, we apply a function of type $\cotokenizer\colon \tokStrings \to \charStrings$ \citep[cf.][]{gastaldi2024foundations}. 
For the purposes of this paper, we assume this pair of functions satisfy:\looseness=-1	

\begin{figure}[t!]
	\centering
	\includegraphics[width=\linewidth]{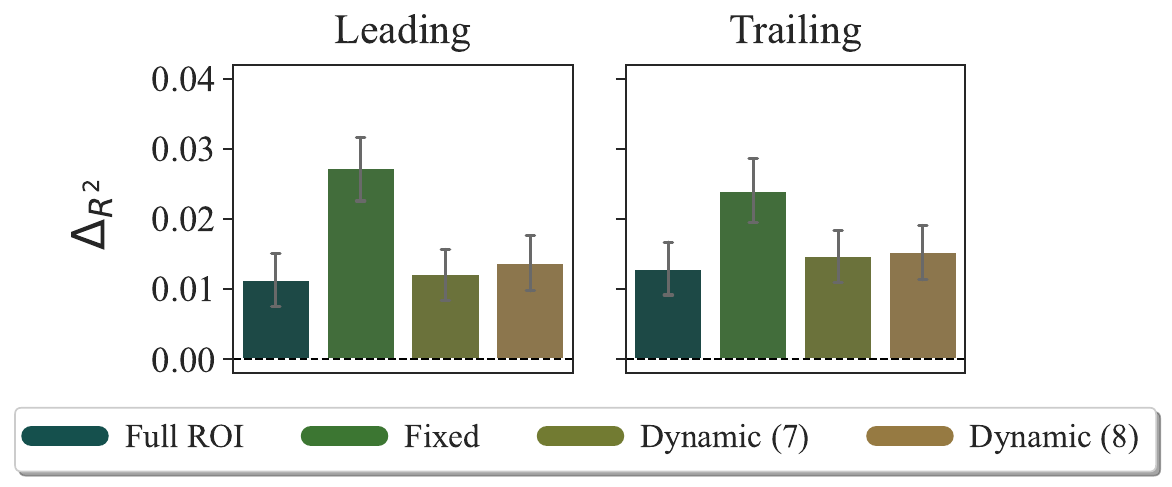}
	\caption{\textbf{Skip Rate}. Predictive power $\drsq$ of an \roi{}'s character-level surprisal, calculated with varying focal areas and two \roi{} types: leading (left) or trailing (right) whitespace. All $\drsq$ scores are significantly above zero ($p<0.001$). Error bars represent 95\% confidence intervals. The black dotted line corresponds to the baseline regressor, including \roi{} length and frequency. The target regressor includes the length, frequency, and full \roi{} surprisal of the previous two \roi{}s.\looseness-1
	}
	\label{fig:skip_rate_celer_rsquared}
	\vspace{-5pt}
\end{figure}

\begin{itemize}[leftmargin=*,topsep=3pt,itemsep=1pt,parsep=1pt]
	\item Exactness: $\forall \charseq \in \charStrings\colon \cotokenizer(\tokenizer(\charseq)) = \charseq$.\footnote{
		But, not necessarily, $\forall \tokseq \in \tokStrings\colon \tokenizer(\cotokenizer(\tokseq)) = \tokseq$.  Thus, we do not require $(\tokenizer, \cotokenizer)$ to form a bijection over $(\charStrings, \tokStrings)$.
	}
	
	\item Multiplicativity: $\cotokenizer(\varepsilon) = \varepsilon$, and $\forall \token_1 \cdots \token_N \in \tokStrings\colon \cotokenizer(\token_1 \cdots \token_N) = \cotokenizer(\token_1) \cdots \cotokenizer(\token_N)$.
\end{itemize}
BPE satisfies both of these properties. 

The probability of a character string $\charseq$ can be computed from a language model over tokens $\ptokens$ using the following marginalization:
\begin{align}
	\pchars(\charseq)
	&=
	\sum_{\tokseq \in \tokStrings}
	\mathbbm{1}\left\{ \charseq = \cotokenizer(\tokseq) \right\}
	\ptokens(\tokseq) .
\end{align}
\noindent Similarly, the prefix probability is given by
\begin{align}
	\preChar(\charseq)
	&=
	\sum_{\tokseq \in \tokStrings}
	\mathbbm{1}\left\{ \charseq \preceq \cotokenizer(\tokseq) \right\}
	\ptokens(\tokseq) .
\end{align}
\citet{vieira-2024-token-to-char-lm} show that $\preChar(\charseq)$ can be computed with a finite summation:
\begin{align}
	\preChar(\charseq)
	&= 
	\sum_{\tokseq \in \cover(\charseq)}
	\preTok(\tokseq),
\end{align}
where $\preTok$ is the prefix probability of $\ptokens$, calculated as in \cref{eq:prefix-prob}, and
the \defn{prefix-cover} $\cover$ is defined as
\begin{align}\label{eq:covering-implication}
	\cover(\charseq) \defeq & \textbf{ if } \charseq = \varepsilon\colon \{\varepsilon\}  \\
	& \textbf{ else}\colon \{ 
	\tokseq' \!\c\! \token \in \tokAlphabet^{\!+} \mid \cotokenizer(\tokseq')  \!\prec\! \charseq \!\preceq\! \cotokenizer(\tokseq' \!\c\! \token) \} .
	\nonumber
\end{align}
Unfortunately, $|\cover(\charseq)|$ can be exponential in $|\charseq|$; thus, we use the beam summing algorithm proposed by \citet{vieira-2024-token-to-char-lm} as a practical approximation algorithm.\footnote{
	See \cref{sec:spurious-ambiguity} for more details.\looseness=-1}  Lastly, to compute the character-level conditional distribution, we use \cref{eq:ratio}, albeit with our approximation to $\preChar(\charseq)$.\looseness=-1

\begin{figure*}
	\centering
	\includegraphics[width=\linewidth]{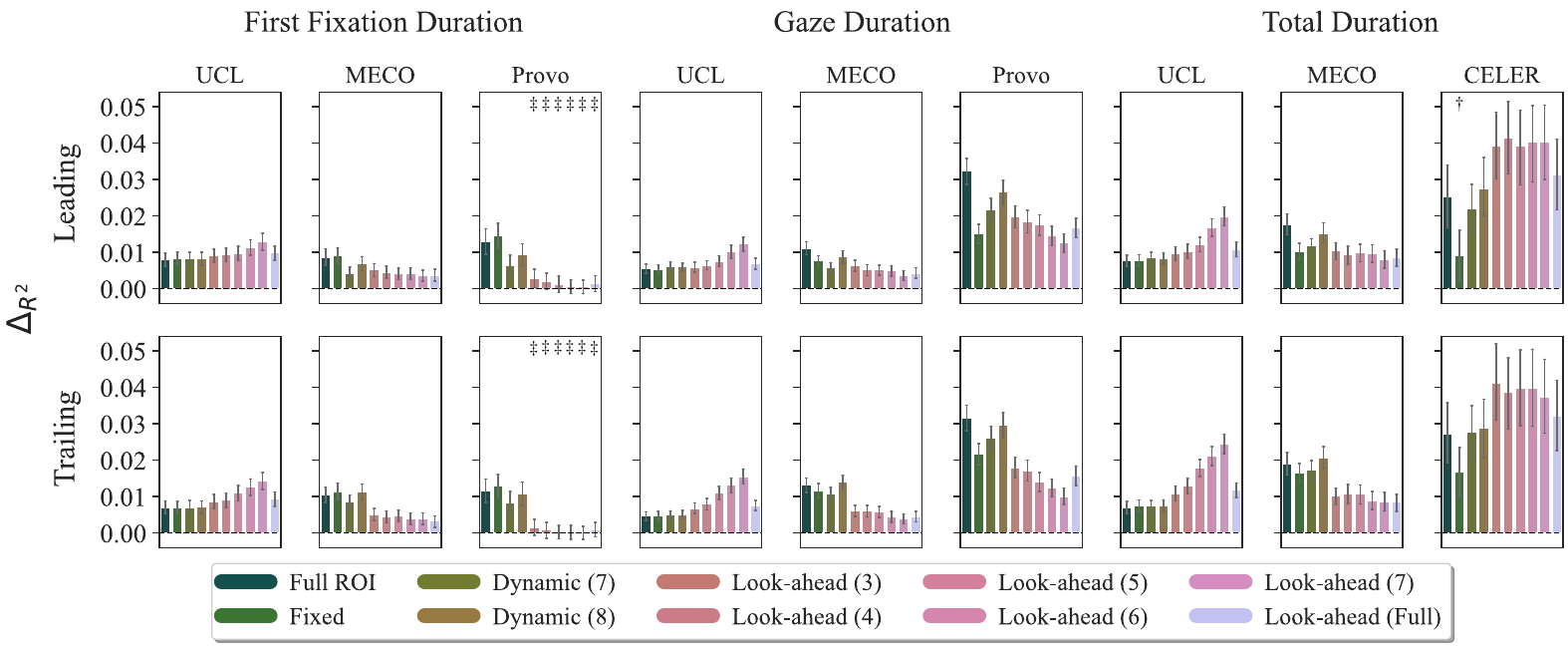}
	\caption{\textbf{First Fixation, Gaze, and Total Duration}. Predictive power $\Delta_{R^2}$ of an \roi{}'s character-level surprisal, calculated with varying focal areas and according to two \roi{} types: with leading (top) or trailing (bottom) whitespace. All $\drsq$ scores are significantly above zero with $p<0.001$, unless marked with $\dag$ ($p<0.01$) or $\ddag$ ($p\geq0.01$). Error bars represent 95\% confidence intervals. The black dotted line represents the baseline regressor, including \roi{} length and frequency. The target regressor includes the length, frequency, and full \roi{} surprisal of the previous two \roi{}s to account for spillover effects.\looseness=-1}
	\label{fig:all_datasets_rsquared_by_focal_area}
	\vspace{-12pt}
\end{figure*}
\section{Predictive Power of Focal Areas}\label{sec:results}

Our experimental design is discussed in \Cref{sec:experimental-design}. We consider \roi{}s with leading and trailing whitespaces and experiment with the focal areas described in \Cref{sec:selecting-focal-areas}. Additional results may be found in \cref{sec:additional-experimental-results}.\looseness=-1

\paragraph{Skip rate.}
The results for skip rates in the CELER dataset \cite{celer2022} are shown in \cref{fig:skip_rate_celer_rsquared} with predictive power expressed as the difference in $R^2$ between target and baseline regressors.
The trends observed are consistent across different \roi{} types. 
Among the predictors examined, the surprisal of the fixed-size focal area, which corresponds to the first three characters of an \roi{}, emerged as the strongest predictor of skipping behavior, with a $\drsq$ significantly higher than all other predictors ($p < 0.001$).
The surprisal of the dynamically sized focal area with a word identification span of 8 characters is the second-best predictor, followed by that of the dynamically sized focal area with a word identification span of 7 characters. 
The surprisal of the full \roi{} is the weakest predictor, with a $\drsq$ approximately two times lower than that of the fixed-size focal area.
These results are consistent with findings that English readers process information to the right of the currently fixated \roi{} collected with human subjects \cite[see, however,][]{heilbron-2023-lexical}.
In particular, they provide new evidence that upcoming \roi{}s are skipped because they are partially read rather than filled in from contextual cues \cite{McConkie-rayner-1975-span,rayner-1975-perceptual,rayner1982availability}, and that the first three characters of the upcoming \roi{} have a special status \cite{lima-inhoff-1985-lexical,lima-1987-morphological}.
Moreover, the fact that the predictive power of the surprisal of fixed-size focal areas is highest for \roi{}s with leading whitespace (13\% higher than fixed-size focal areas with trailing whitespace; $p<0.01$) suggests a dual model of skipping decisions.
On the one hand, parafoveal preview provides cues for lexical identification \cite{inhoff-1989-lexical,inhoff-1990-integrating,inhoff-tousman-1990-lexical} and on the other, word boundary information to the right of the currently fixated \roi{} is used for saccade planning \cite{rayner-pollatsek-1981-eye,pollatsek-1982-eye}.\looseness=-1

\paragraph{First fixation, gaze, and total duration.}
The results for the UCL, Provo, MECO, and CELER datasets are shown in \cref{fig:all_datasets_rsquared_by_focal_area}, with predictive power expressed as the difference in $R^2$ between target and baseline regressors.
Look-ahead focal areas significantly improve reading time predictions on the UCL dataset (first fixation duration, gaze duration, and total duration) and the CELER dataset (total duration), compared to using focal areas over the entire \roi{}.
In the Provo and MECO datasets, the predictive power of surprisal from the fixed-size and dynamically sized focal areas is comparable to that of the full \roi{}'s surprisal. 
For first fixation duration, the fixed-size focal area surprisal emerges as the best predictor, alongside the full \roi{}'s surprisal. 
For gaze and total duration, the surprisal from the dynamically sized focal area, with a rightward word identification span of 8 characters, is on par with the full \roi{}'s surprisal; their relative ranking varies across the two types of \roi{}. 
These findings align with psycholinguistic evidence that the perceptual span of English readers extends to the right of the current fixation \cite{McConkie-rayner-1975-span,denbuurman-etal-1981-eye,underwood-mcconkie-1985-perceptual}.
The variability in results across datasets reinforces the view that \roi{} and focal area definitions are an empirical matter and points to a nuanced perspective on how character-level information influences reading behavior.
The strong predictive power of look-ahead focal areas, which span both the current and upcoming \roi{}, indicates that parafoveal information affects saccade latency for the \emph{currently} fixated \roi{}---an effect which could be connected to the assessment of longer-horizon prediction errors \cite{giulianelli-etal-2024-incremental} or to the front-loading of integration costs that may occur when previewing the upcoming region \cite{gibson2001dependency,futrell2020lossy}.
However, the predictive power of fixed-size and dynamically sized focal areas, which span only part of the \roi{} and model saccade latency based on the preceding fixation, suggests that parafoveal information is also used to preprocess \textit{upcoming} \roi{}s.\looseness=-1

\section{Conclusion}
We treat the role of tokenization in psycholinguistics.
We recommend predictors be derived from character-level surprisal, allowing the modeler to explore a wider range of useful predictors.\looseness=-1 

\section*{Limitations}
Our analyses are conducted exclusively on English stimuli and measurements collected from L1 English readers.
Additionally, we focus solely on eye-tracking data, as it is more natural to conceptualize focal areas in this context. 
We do not analyze self-paced reading, where the challenges are likely even more complex due to the variability in how the method is applied. 
For example, some studies use a moving-window paradigm in which words are masked by dashes \citep{just1982paradigms}, preserving whitespace information, whereas others rely on centered presentation \citep{aaronson-1976-performance}, which omits whitespace information by design.
Further complications may arise from differences between word-by-word and chunked presentation \citep[see, e.g.,][]{tremblay-etal-2011-processing}, where both single-word and multi-word \roi{}s may be considered, as well as from paradigms presenting multiple alternative \roi{}s at a time \citep{forster2009maze,boyce2023maze}. 
How these variations interact with surprisal predictors remains poorly understood, and future work is necessary to model self-paced reading data more comprehensively using focal area predictors.\looseness=-1

Other limitations of our approach lie in the modeling assumptions made about the relationship between surprisal and reading behaviors.
While we employ linear modeling based on established evidence that the relationship between surprisal and reading time is linear \cite{smith2008optimal,smith2013-log-reading-time,wilcox-2023-testing,shainetal24}, this relationship has not yet been determined for skip rates. 
To our knowledge, no studies have examined skip rates using our focal area predictors, and the functional relationship between surprisal and skip rates remains to be determined. 
Future research should investigate skip rates with modeling approaches capable of capturing non-linear relationships, such as generalized additive models \citep[GAMs;][]{wood-2004-stable,wood2017generalized}.
Finally, we do not account for individual differences between participants in our analysis, which could be more accurately modeled using mixed-effects models \citep{gelmanbda04}.\looseness=-1

\section*{Acknowledgements}
We thank Shayne Sloggett for a lengthy discussion about the state of the self-paced reading paradigm as well as his sage insights into the eye-tracking literature.
We also thank Patrick Haller, Ethan Gotlieb Wilcox, Yahya Emara, Ekaterina Vylomova, and Eleanor Chodroff for useful discussions about the psycholinguistic content of the paper, and Clemente Pasti, Robin Shing Moon Chan, Zeerak Talat, Anej Svete, and Vésteinn Snæbjarnarson for help with copy editing. 
Mario Giulianelli was supported by an ETH Zurich Postdoctoral Fellowship.

\bibliography{main}

\appendix 
\onecolumn

\section{A Note on Spurious Ambiguity}\label{sec:spurious-ambiguity}
The potentially exponential size of $|\cover(\charseq)|$ stems from \defn{spurious ambiguity}, i.e., when character strings in $\kleene{\alphabet}$ correspond to more than one token string in $\kleene{\tokAlphabet}$ \citep[cf.][]{gastaldi2024foundations}.
For example, due to its BPE-based tokenizer, \textsc{GPT-2} can generate the character string \charfont{footprint} using its canonical tokenization \tokenfont{foot print},
but it can also generate hundreds of non-canonical tokenizations of the same word, e.g., \tokenfont{foot pr in t} and \tokenfont{f o o t p r i n t}, all of which have to be accounted for when marginalizing a token-level language model to a character-level language model. 
In the case that there are many such tokenizations, it is \#P-hard to compute character-level surprisal exactly \citep{geh2024signaltokenizationspace}, justifying our reliance on an approximate beam summing algorithm.

\section{Experimental Design}\label{sec:experimental-design}
Our experiments investigate the predictive power of the surprisal of different focal areas for reading time measurements across four eye-tracking datasets, presented in \cref{sec:psyling-data}.
We consider \roi{}s including either a leading or a trailing whitespace, estimate the surprisal for the focal areas of \roi{}s using a parameterized language model (\cref{sec:exp-setup-lm}), and then fit a statistical model to the average
measurements\footnote{We average an \roi{}'s measurements across participants, following a standard procedure used extensively in prior work \citep[][\textit{inter alia}]{smith2013-log-reading-time,wilcox2020, meister-etal-2021-revisiting,devarda-etal-2023,wilcox-2023-testing}. 
	See \citet{smith2013-log-reading-time} for experiments verifying that this leads to the same linear surprisal effects in eye-tracked reading time datasets.
} 
of human reading behavior using surprisal estimates as predictors, as outlined in \cref{sec:linear-modeling}.
We describe each component of our experimental setup in the following sections.
\looseness=-1

\subsection{Data}
\label{sec:psyling-data}
We analyze four datasets annotated with reading time measurements collected in eye-tracking experiments with human particpants:
UCL \cite{frank2013reading}, Provo \cite{provo-2018}, MECO \cite{siegelman2022expanding}, and CELER \citep{celer2022}.
For MECO and CELER, both multilingual datasets, we include only data from English stimuli and participants with English as their first language (L1).
These eye-tracking datasets provide several \roi{}-based measurements of reading times. 
Our study focuses on four specific measurements \cite{rayner1998eye}: first fixation duration, gaze duration, total duration, and skip rate.\looseness-1

\paragraph{UCL \citep{frank2013reading}.}
The UCL Corpus of eye-tracked reading times contains 205 stimuli extracted from three English novels. This datasets attempts to serve as a gold standard for evaluating computational psycholinguistic models of English sentence comprehension. It addresses limitations of previous datasets by using independent sentences that can be understood without extensive context or extra-linguistic knowledge. The corpus includes data from 43 subjects who were recruited from the University College London subject pool. Eye movements were recorded using a head-mounted EyeLink II eyetracker with a 500 Hz sampling rate.
Stimuli range from 5 to 15 words, for a total of 1726 \roi{}s; measurements for the first \roi{} in a stimulus are omitted. 
We analyze first fixation duration, gaze duration, and total (right-bounded fixation) duration, and exclude go-past time, a measurement that includes the time spent by the reader in regressions to previous words.

\paragraph{Provo \cite{provo-2018}.} This corpus consists of 136 sentences of English text from a variety of genres, including online news articles, popular science, and public-domain works of fiction. 
These sentences were presented as part of 55 short passages, with an average length of 50 words and 2.5 sentences per passage.
Eye movement data was collected from 84 native English speakers using an SR Research EyeLink 1000 Plus eye-tracker. Participants read the texts for comprehension while their eye movements were recorded.
The Provo corpus was designed to facilitate the investigation of predictability effects in reading and offers a more naturalistic distribution of word predictability compared to traditional sentence completion norms.
In this work, we analyze first fixation duration, gaze duration, and skip rate.

\paragraph{MECO \citep{siegelman2022expanding}.}
The Multilingual Eye Movement Corpus (MECO) contains eye-tracking data from L1 speakers (between 29 and 54 per language) for 12 simplified Wikipedia-style articles in 13 languages.
In our analysis, we only include English stimuli and responses from 46 L1 speakers of English. 
We analyze first fixation duration, gaze duration, and total duration for comparability with previous work \citep{wilcox-2023-testing}.

\paragraph{CELER \citep{celer2022}.} The Corpus of Eye Movements in L1 and L2 English Reading (CELER) is a large-scale eye-tracking dataset focused on English reading, consisting of data from 365 participants, including 69 native English speakers (L1) and 296 non-native English speakers (L2). The dataset contains over 320,000 words, with each participant reading 156 newswire sentences from the Wall Street Journal. CELER includes reading time and eye movement data (collected using Eyelink 1000 and Eyelink 1000 Plus eye-trackers) for each sentence. Participants are asked comprehension questions to assess their understanding of the read text. In this paper, we consider only L1 English speakers reading sentences shared across all participants, and discarding sentences unique to a single reader. 
We analyze first fixation duration, gaze duration, total duration, and skip rate.

\subsubsection{Data Preprocessing}
We only apply a simple data filtering step: We skip the first region of every stimulus.
This makes our analysis consistent across datasets (as measurements corresponding to the first region are not always available) and comparable with prior work, which adopts the same procedure \citep{frank2013reading,goodkind2018predictive,devarda-etal-2023}.

\subsection{Language Models}\label{sec:exp-setup-lm}
All experiments are conducted using \textsc{GPT-2} \citep{radford2019language} in its \texttt{small} variant.
Despite its size, \textsc{GPT-2} \texttt{small} has been shown to have greater predictive power for reading time data than larger models \cite{oh2022does,shainetal24}. 
As explained in \cref{sec:marginalizing-out-token-strings}, we use the beam summing algorithm proposed by \citet{vieira-2024-token-to-char-lm} to compute the surprisal of focal areas, setting the beam size to 5.

\subsection{Linear Modeling}\label{sec:linear-modeling}
Given a dataset of \roi{}s extracted from a corpus of psycholinguistic stimuli, our goal is to design a statistical model that explains the measurements associated with each stimulus in terms of surprisal predictors.
We use linear modeling for our regression analyses as previous work has shown the relationship between surprisal and reading time measurements is largely linear \citep{smith2008optimal,smith2013-log-reading-time,wilcox-2023-testing,shainetal24}.
To investigate the predictive power of the surprisal of different focal areas, we employ a 2-by-10 design, with two ways of constructing regions of interest (see \cref{sec:formalizing-roi}) and ten ways of defining a region's focal area (see \cref{sec:focal-areas}).
For regions of interest, we include segmentative sequences with a leading whitespace and segmentative sequences with a trailing whitespace.
As focal areas, we consider the entire region, a fixed-size focal area covering the first three characters of the region, dynamically sized focal areas with a rightward word identification span of 7 and 8 characters, and look-ahead focal areas peeking into the next 3--7 characters as well as spanning over the entire next \roi{}. 
\cref{tab:focal-areas-example} shows the twenty sequences of focal areas that result from this experimental design for an example stimulus.
The next sections describe the analysis procedure in detail, first presenting our metric of predictive power (\cref{sec:linear-modelling-predictive-power}), then the specific predictors used in our linear models (\cref{sec:linear-modelling-baselines} and \cref{sec:linear-modelling-targets}).

\subsubsection{Predictive Power}\label{sec:linear-modelling-predictive-power}
For each combination of \roi{} and focal area type, we compare a \defn{baseline regressor} including well-established predictors of reading behavior (frequency and length) to a \defn{target regressor} which, on top of the baseline predictors, includes the surprisal of the focal area.
To isolate the true predictive power contributed by a target predictor of interest (i.e., the surprisal of an \roi{}'s focal area) from that of baseline predictors (i.e., an \roi{}'s length and frequency), we inspect the difference in $\rsq$ assigned to a held-out set by the baseline regressor and the target regressor, which we denote as $\drsq$.
We estimate $\drsq$ via 10-fold cross-validation, iterating over 10 random seeds. We fit the regressor on 9 data folds at a time by finding the coefficients that minimize the residual sum of squares, and then measure the regressor's $\rsq$ on the 10\textsuperscript{th} fold to evaluate its fit.
As our final measure of \defn{predictive power}, we report the average $\drsq$ across folds and random seeds, with 95\% confidence intervals.
To assess the statistical significance of a target predictor's $\drsq$, we run paired permutation tests with the cross-validation results;\footnote{
	We use the implementation provided by the \texttt{SciPy} library under \texttt{\href{https://docs.scipy.org/doc/scipy/reference/generated/scipy.stats.permutation_test.html}{scipy.stats.permutation\_test}}. 
}
see \citet{giulianelli-etal-2024-generalized} for a detailed description of these significance tests.
Finally, for comparison with prior work \cite[][\textit{inter alia}]{goodkind2018predictive,wilcox2020,wilcox-2023-testing}, we also calculate the average per-\roi{} difference in log-likelihood $\dllh$ of the test set between the target and baseline regressor, and report $\dllh$ as an additional metric of predictive power (see \cref{sec:additional-experimental-results}). 

\subsubsection{Baseline Predictors} \label{sec:linear-modelling-baselines}
We consider two baseline predictors: the length of an \roi{} measured in characters and the logarithm of the \roi{}'s frequency, obtained using the \texttt{wordfreq} software.\footnote{
	We use the Zipf frequency \citep{brysbaert2012adding}, i.e., the base-10 logarithm of the number of times an \roi{} (with whitespaces removed) appears per billion words.
}
Both length and frequency are well-established predictors of reading times \cite{rayner1998eye}. 
The impact of length is fairly intuitive, and the effectiveness of frequency as a predictor has been demonstrated in numerous studies \cite[][\textit{inter alia}]{inhoff-reyner-1985-parafoveal,rayner-duffy-1986-lexical,hyona-olson-1995-eye}.\looseness-1

\subsubsection{Target Predictors}\label{sec:linear-modelling-targets}
As the target predictor for a given \roi{}'s measurements, we use the surprisal of the \roi{}'s focal area, calculated as described in \cref{sec:exp-setup-lm}.
Finally, to account for spillover effects, we also include in the target regressor the length, log-frequency, and full \roi{} surprisal of the previous two \roi{}s \citep{just1982paradigms,frank2013reading}.
The surprisal of the previous two \roi{}s is calculated according to the \roi{} definition (with leading or trailing whitespace) of the main target predictor.
This makes predictive power scores comparable across \roi{}s and focal areas.

\section{Additional Experimental Results}\label{sec:additional-experimental-results}
\begin{figure}[h!]
	\centering
	\includegraphics[width=0.5\linewidth]{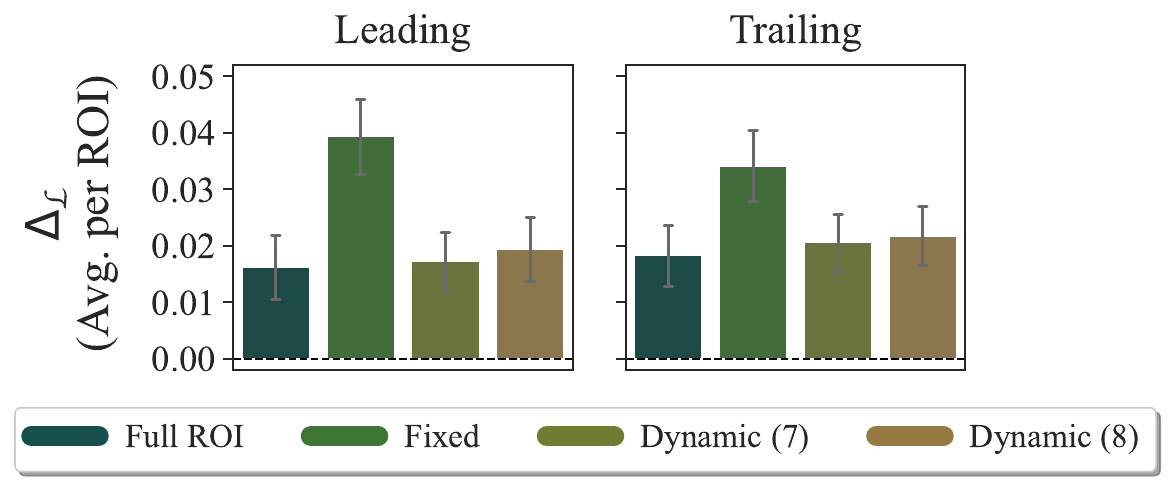}
	\caption{\textbf{Skip Rate (CELER)}. Predictive power $\dllh$ of an \roi{}'s character-level surprisal, calculated with varying focal areas and two \roi{} types: leading (left) or trailing (right) whitespace. All $\dllh$ scores are significantly above zero ($p<0.001$). Error bars represent 95\% confidence intervals. The black dotted line corresponds to the baseline regressor, including \roi{} length and frequency. The target regressor includes the length, frequency, and full \roi{} surprisal of the previous two \roi{}s.\looseness-1
	}
	\label{fig:skip_rate_celer_loglik}
	\vspace{-10pt}
\end{figure}

\begin{figure}[h!]
	\centering
	\begin{subfigure}[b]{0.49\textwidth}
		\centering
		\includegraphics[width=\textwidth]{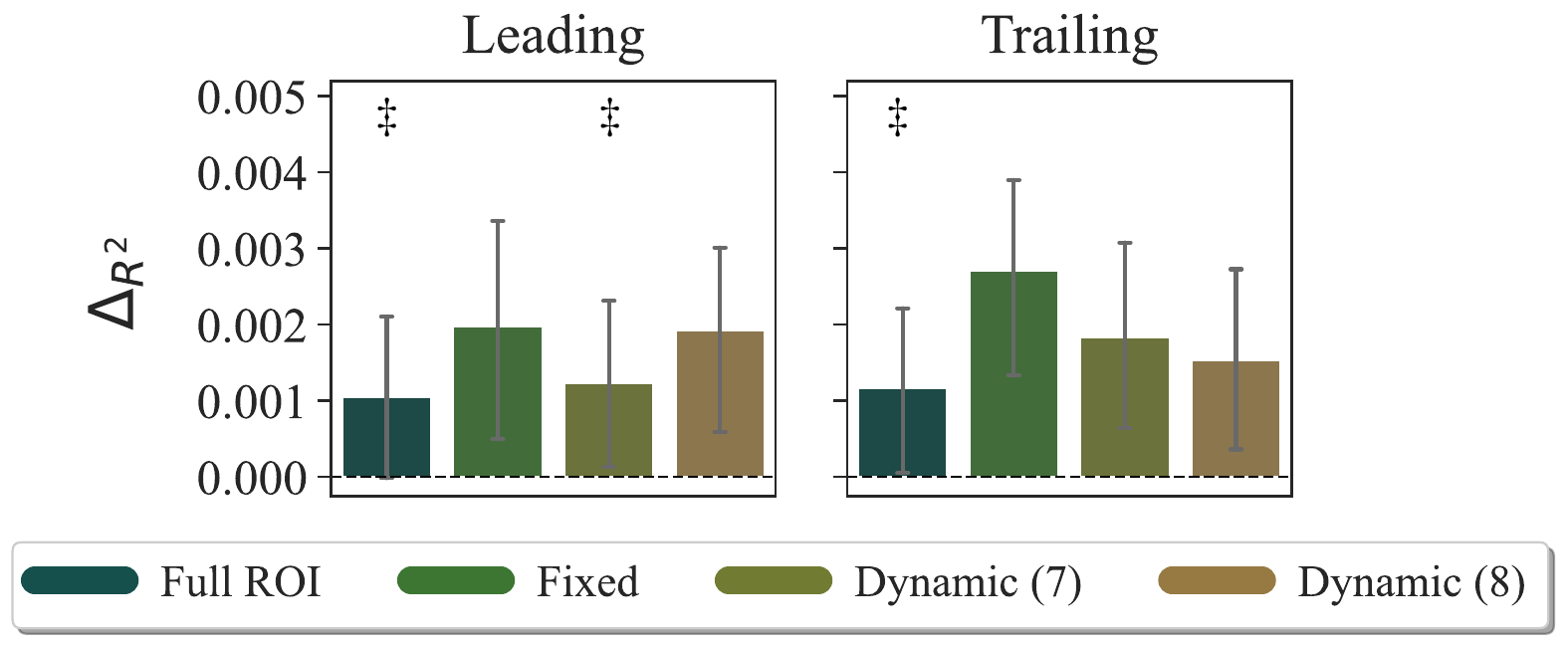}
		\label{fig:skip_rate_provo_rsquared}
		\vspace{-1em}
		\caption{}
	\end{subfigure}
	\hfill
	\begin{subfigure}[b]{0.49\textwidth}
		\centering
		\includegraphics[width=\textwidth]{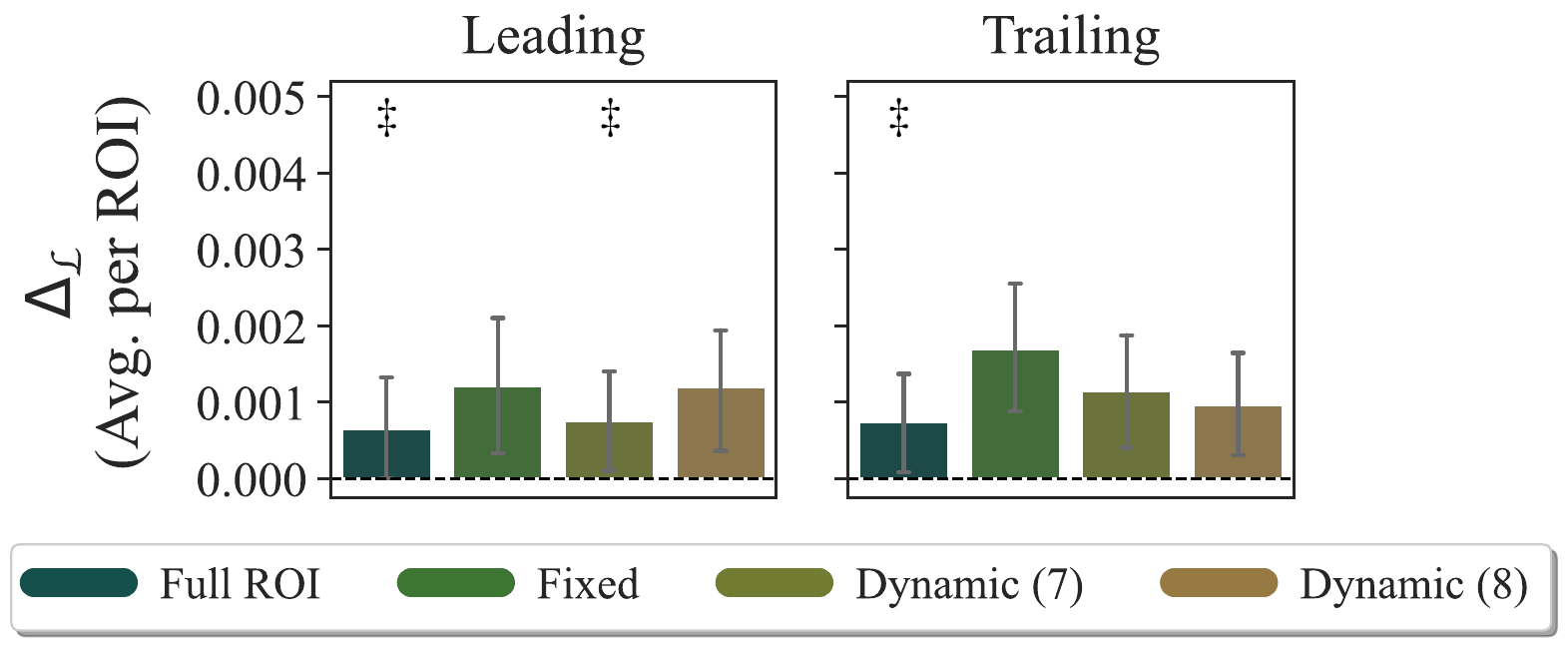}
		\label{fig:skip_rate_provo_loglik}
		\vspace{-1em}
		\caption{}
	\end{subfigure}
	\caption{\textbf{Skip Rate (Provo)}. Predictive power (subfigure (a): $\drsq$; subfigure (b): $\dllh$) of an \roi{}'s character-level surprisal, calculated with varying focal areas and two \roi{} types: leading or trailing whitespace (respectively left and right within each subfigure). All $\drsq$ and $\dllh$ scores are significantly above zero with $p<0.01$, unless marked with $\ddag$ ($p\geq0.01$).
		Error bars represent 95\% confidence intervals. The black dotted line corresponds to the baseline regressor, including \roi{} length and frequency. The target regressor includes the length, frequency, and full \roi{} surprisal of the previous two \roi{}s.\looseness-1}
	\label{fig:skip_rate_provo}
\end{figure}

\begin{figure*}[h!]
	\centering
	\includegraphics[width=\linewidth]{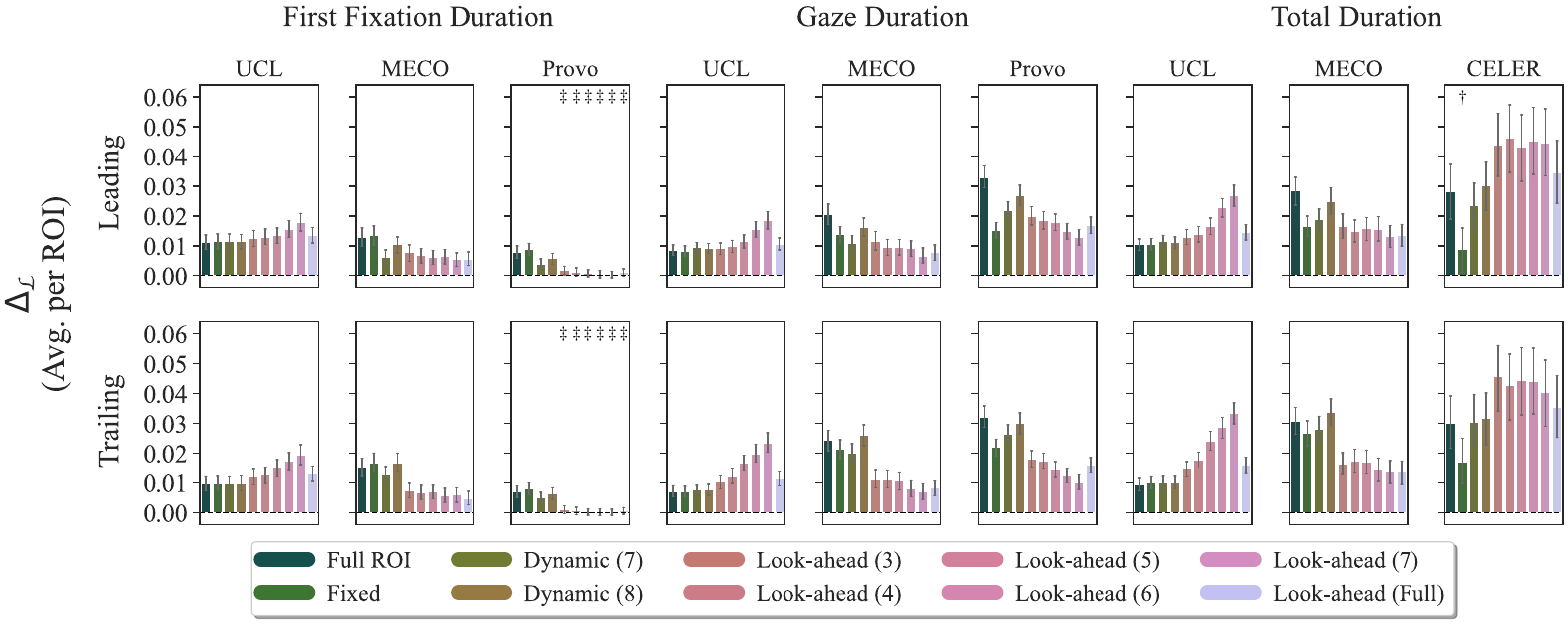}
	\caption{\textbf{First Fixation, Gaze, and Total Duration}. Predictive power $\dllh$ of an \roi{}'s character-level surprisal, calculated with varying focal areas and according to two \roi{} types: with leading (top) or trailing (bottom) whitespace. All $\dllh$ scores are significantly above zero with $p<0.001$, unless marked with $\dag$ ($p<0.01$) or $\ddag$ ($p\geq0.01$). Error bars represent 95\% confidence intervals. The black dotted line represents the baseline regressor, including \roi{} length and frequency. The target regressor includes the length, frequency, and full \roi{} surprisal of the previous two \roi{}s to account for spillover effects.\looseness=-1}
	\label{fig:all_datasets_loglik_by_focal_area}
	\vspace{-7.5pt}
\end{figure*}

Our experiments investigate the predictive power of focal areas for eye-tracked measurements of reading behavior.
In \cref{sec:results} of the main paper, we report results on skip rate (CELER), first fixation duration (UCL, MECO, and Provo), gaze duration (UCL, MECO, and Provo), total duration (UCL, MECO, and CELER), using $\drsq$ as our metric of predictive power.
On CELER, we find no significant predictors of first fixation duration and gaze duration.
Here, we report further results on skip rate in CELER, using $\dllh$ as a metric of predictive power (\cref{fig:skip_rate_celer_loglik}) and in Provo (\cref{fig:skip_rate_provo}, both with $\drsq$ and $\dllh$).
The latter follow the same trends as skip rate in CELER albeit with lower predictive power. 
Finally, \cref{fig:all_datasets_loglik_by_focal_area} shows results on first fixation duration, gaze duration, and total duration on the same datasets as in \cref{sec:results} but using $\dllh$ as a metric of predictive power. 
\vfill
\vfill

\end{document}

%% file: macros.tex
\newcommand{\mymacro}[1]{#1}

\newcommand{\defn}[1]{\textbf{#1}}
\newcommand{\kleene}[1]{{\mymacro{#1^*}}}
\newcommand{\kleeneplus}[1]{{\mymacro{#1^+}}}

\newcommand{\eos}{{\mymacro{\textsc{eos}}}}
\newcommand{\defeq}{\mathrel{\stackrel{\textnormal{\tiny def}}{=}}}

\newcommand{\roi}{\textsc{ROI}}
\newcommand{\region}[1]{\mymacro{\boldsymbol{\rho}_{#1}}}

\newcommand{\datum}{\mymacro{\psi}}
\newcommand{\surprisal}{\mymacro{\iota}}

\newcommand{\prefViewLoc}{\mymacro{v}}
\newcommand{\wordIdSpan}{\mymacro{s}}

\newcommand{\rsq}{{\mymacro{R^2}}}
\newcommand{\drsq}{{\mymacro{\Delta_{\rsq}}}}
\newcommand{\llh}{{\mymacro{\mathcal{L}}}}
\newcommand{\dllh}{{\mymacro{\Delta_{\llh}}}}

\newcommand{\charAlphabet}{\mymacro{\Sigma}}

\newcommand{\stimulus}{\mymacro{\boldsymbol{\sigma}}}
\newcommand{\charSym}[1]{\mymacro{\sigma_{#1}}}

\newcommand{\pLM}{\mymacro{p}}

\newcommand{\pLMprefix}{\mymacro{\overrightarrow{\pLM}}}

\newcommand{\pHum}{\mymacro{\pLM_{\mathrm{H}}}}

\newcommand{\tokenfont}[1]{{\color{orange!60!black}\texttt{{\small#1}}}}
\newcommand{\charfont}[1]{{\color{blue!80!black}\texttt{{\small#1}}}}
\newcommand{\focalareafont}[1]{{\color{green!50!black}\texttt{{\small\underline{#1}}}}}

\newcommand{\charStrings}[0]{\kleene{\alphabet}}

\newcommand{\paroutline}[3][false]{%
    \ifnum\pdfstrcmp{#1}{true}=0
        #3%
    \else
        [\textit{\textcolor{DiverseMagenta}{#2}}] \textcolor{AccentBlue}{#3}%
    \fi
}

\newcommand{\cover}{{\mymacro{\mathcal{C}}}}

\newcommand{\alphabet}{{\mymacro{ \Sigma}}}

\newcommand{\tokAlphabet}{{\mymacro{\Delta}}}

\renewcommand{\c}[0]{\mymacro{\cdot}}

\newcommand{\tokenizer}{\ensuremath{\mymacro \tau}\xspace}
\newcommand{\cotokenizer}{\ensuremath{\mymacro \kappa}\xspace}

\newcommand{\R}{\mathbb{R}}
\newcommand{\tokseq}{\ensuremath{\mymacro{\boldsymbol{\delta}}}\xspace}
\newcommand{\token}{\mymacro{\delta}}
\newcommand{\tokStrings}[0]{\kleene{\tokAlphabet}}
\newcommand{\ptokens}{{\mymacro{p_{\tokAlphabet}}}}

\newcommand{\charseq}{\ensuremath{\mymacro{\boldsymbol{\sigma}}}\xspace}

\newcommand{\pchars}{{\mymacro{p_{\alphabet}}}}

\newcommand{\preTok}{\mymacro{\overrightarrow{p_{\tokAlphabet}}}}
\newcommand{\preChar}{\mymacro{\overrightarrow{p_{\alphabet}}}}

\usepackage{linguex}

\usepackage{nameref}
\usepackage[capitalize,noabbrev]{cleveref}

\crefformat{section}{\S#2#1#3}
\crefformat{subsection}{\S#2#1#3}
\crefformat{subsubsection}{\S#2#1#3}
\crefformat{paragraph}{\P#2#1#3}
\crefformat{subparagraph}{\P#2#1#3}
\crefmultiformat{section}{\S#2#1#3}{ and~\S#2#1#3}{, \S#2#1#3}{, and~\S#2#1#3}
\crefmultiformat{subsection}{\S#2#1#3}{ and~\S#2#1#3}{, \S#2#1#3}{, and~\S#2#1#3}
\crefmultiformat{subsubsection}{\S#2#1#3}{ and~\S#2#1#3}{, \S#2#1#3}{, and~\S#2#1#3}
\crefmultiformat{paragraph}{\P\P#2#1#3}{ and~#2#1#3}{, #2#1#3}{, and~#2#1#3}
\crefmultiformat{subparagraph}{\P\P#2#1#3}{ and~#2#1#3}{, #2#1#3}{, and~#2#1#3}
\crefrangeformat{section}{\mbox{\S\S#3#1#4--#5#2#6}}
\crefrangeformat{subsection}{\mbox{\S\S#3#1#4--#5#2#6}}
\crefrangeformat{subsubsection}{\mbox{\S\S#3#1#4--#5#2#6}}
\crefrangeformat{paragraph}{\mbox{\P\P#3#1#4--#5#2#6}}
\crefrangeformat{subparagraph}{\mbox{\P\P#3#1#4--#5#2#6}}
\crefname{part}{Part}{Parts}
\Crefname{part}{Part}{Parts}
\crefname{chapter}{Ch.}{Ch.}
\Crefname{chapter}{Ch.}{Ch.}
\crefname{footnote}{Fn.}{Fn.}
\Crefname{footnote}{Fn.}{Fn.}
\crefname{figure}{Figure}{Figures}
\crefname{table}{Table}{Tables}
\crefname{subfigure}{Figure}{Figures}
\Crefname{subfigure}{Figure}{Figures}
\crefname{appsec}{Appendix}{Appendices}
\Crefname{appsec}{Appendix}{Appendices}
\crefname{algocf}{Algorithm}{Algorithms}
\Crefname{algocf}{Algorithm}{Algorithms}

\crefname{ExNo}{example}{examples}
\Crefname{ExNo}{Example}{Examples}
\crefname{SubExNo}{example}{examples}
\Crefname{SubExNo}{Example}{Examples}
\crefname{SubSubExNo}{example}{examples}
\Crefname{SubSubExNo}{Example}{Examples}
\crefformat{ExNo}{(#2#1#3)}
\crefformat{SubExNo}{(#2#1#3)}
\crefformat{SubSubExNo}{(#2#1#3)}
\crefrangeformat{ExNo}{\mbox{(#3#1#4--#5#2#6)}}
\crefrangeformat{SubExNo}{\mbox{(#3#1#4--#5#2#6)}}
\crefrangeformat{SubSubExNo}{\mbox{(#3#1#4--#5#2#6)}}
\crefmultiformat{ExNo}{(#2#1#3}{, #2#1#3)}{, #2#1#3}{, #2#1#3)}
\crefmultiformat{SubExNo}{(#2#1#3}{, #2#1#3)}{, #2#1#3}{, #2#1#3)}
\crefmultiformat{SubSubExNo}{(#2#1#3}{, #2#1#3)}{, #2#1#3}{, #2#1#3)}
\crefrangemultiformat{ExNo}{(#3#1#4--#5#2#6}{, #3#1#4--#5#2#6)}{, #3#1#4--#5#2#6}{, #3#1#4--#5#2#6)}
\crefrangemultiformat{SubExNo}{(#3#1#4--#5#2#6}{, #3#1#4--#5#2#6)}{, #3#1#4--#5#2#6}{, #3#1#4--#5#2#6)}
\crefrangemultiformat{SubSubExNo}{(#3#1#4--#5#2#6}{, #3#1#4--#5#2#6)}{, #3#1#4--#5#2#6}{, #3#1#4--#5#2#6)}

\ifx\creflastconjunction\undefined%
\newcommand{\creflastconjunction}{, and\nobreakspace} %
\else%
\renewcommand{\creflastconjunction}{, and\nobreakspace} %
\fi%

\newcommand*{\Fullref}[1]{\hyperref[{#1}]{\Cref*{#1}: \nameref*{#1}}}
\newcommand*{\fullref}[1]{\hyperref[{#1}]{\cref*{#1}: \nameref{#1}}}

\crefname{section}{\S}{\S\S}
\crefname{table}{Tab.}{Tab.}
\crefname{figure}{Fig.}{Figs.}
\crefname{algorithm}{Alg.}{}
\crefname{equation}{Eq.}{Eq.}
\crefname{appendix}{App.}{}
\crefname{theorem}{Theorem}{}
\crefname{prop}{Proposition}{}
\crefname{definition}{Def.}{}
\crefname{cor}{Corollary}{}
\newtheorem{remark}{Remark}
\crefname{observation}{Observation}{}
\crefname{assumption}{Assumption}{}
\crefname{hyp}{Hyp.}{Hypotheses}
\crefformat{section}{\S#2#1#3}
\crefname{namedtheorem}{Hyp.}{Hypotheses}